\documentclass[10pt,twocolumn,letterpaper]{article}

\usepackage{cvpr}
\usepackage{times}
\usepackage{epsfig}
\usepackage{graphicx}
\usepackage{amsmath}
\usepackage{amssymb}

\usepackage{multirow}
\usepackage{mathrsfs}
\usepackage{enumitem}
\usepackage{makecell}
\usepackage{tabulary}
\usepackage{pifont}
\usepackage{subfigure}
\usepackage{bm}

\usepackage[breaklinks=true,bookmarks=false]{hyperref}

\cvprfinalcopy 

\def\cvprPaperID{9683} 

\ifcvprfinal\fi
\begin{document}

\title{Auto-captions on GIF: A Large-scale Video-sentence Dataset for\\ Vision-language Pre-training}

\author{Yingwei Pan, Yehao Li, Jianjie Luo, Jun Xu, Ting Yao, and Tao Mei \\
{\normalsize\centering JD AI Research, Beijing, China}\\
{\tt\small \{panyw.ustc, yehaoli.sysu, jianjieluo.sysu\}@gmail.com}
}

\maketitle
\thispagestyle{empty}


\begin{abstract}
In this work, we present \textbf{Auto-captions on GIF}, which is a new large-scale pre-training dataset for generic video understanding. All video-sentence pairs are created by \textbf{automatically} extracting and filtering video caption annotations from billions of web pages. Auto-captions on GIF dataset can be utilized to pre-train the generic feature representation or encoder-decoder structure for video captioning, and other downstream tasks (e.g., sentence localization in videos, video question answering, etc.) as well. We present a detailed analysis of Auto-captions on GIF dataset in comparison to existing video-sentence datasets. We also provide an evaluation of a Transformer-based encoder-decoder structure for vision-language pre-training, which is further adapted to video captioning downstream task and yields the compelling generalizability on MSR-VTT. The dataset is available at \url{http://www.auto-video-captions.top/2020/dataset}.
\end{abstract}

\section{Introduction}
Vision-language pre-training has been an emerging and fast-developing research topic in image domain \cite{lu2019vilbert,su2019vl,tan2019lxmert,zhou2019unified}, which transfers multi-modal knowledge from rich-resource pre-training task to limited-resource downstream tasks (e.g., visual question answering \cite{anderson2017bottom,antol2015vqa}, cross-modal retrieval \cite{li2019learning,pan2014click,yao2015learning}, image captioning \cite{li2019novel,yao2017novel,yao2018exploring,yao2019hierarchy,yao2017boosting}, and image paragraph generation \cite{wang2019paragraph}). Nevertheless, the pre-training of generic feature or structure for video understanding is seldom explored and remains challenging.
This is in part due to the simplicity of current video-sentence benchmarks, which mostly focus on specific fine-grained domains with limited videos (e.g., cooking scenario \cite{das2013thousand:CVPR13,regneri:TACL2013,Rohrbach:ICCV13} and movie domain \cite{Rohrbach:CVPR15,Torabi15}). Furthermore, the human annotations (e.g., video-sentence pairs) are resourcefully expensive and thus cannot be scaled up.

In this paper, we present the Auto-captions on GIF dataset, which is a new large-scale video-sentence benchmark for vision-language pre-training, to pursue the generic video understanding. This is achieved by automatically extracting, filtering, and refining raw descriptions from the Alt-text HTML attribute of web GIF videos in billions of web pages. In particular, an automatic pipeline is devised to extract, filter, and refine the raw video-sentence pairs, leading to the current version of Auto-captions on GIF with 164,378 video-sentence pairs.

\begin{figure}[!tb]
\vspace{-0.0in}
\centering {\includegraphics[width=1\linewidth]{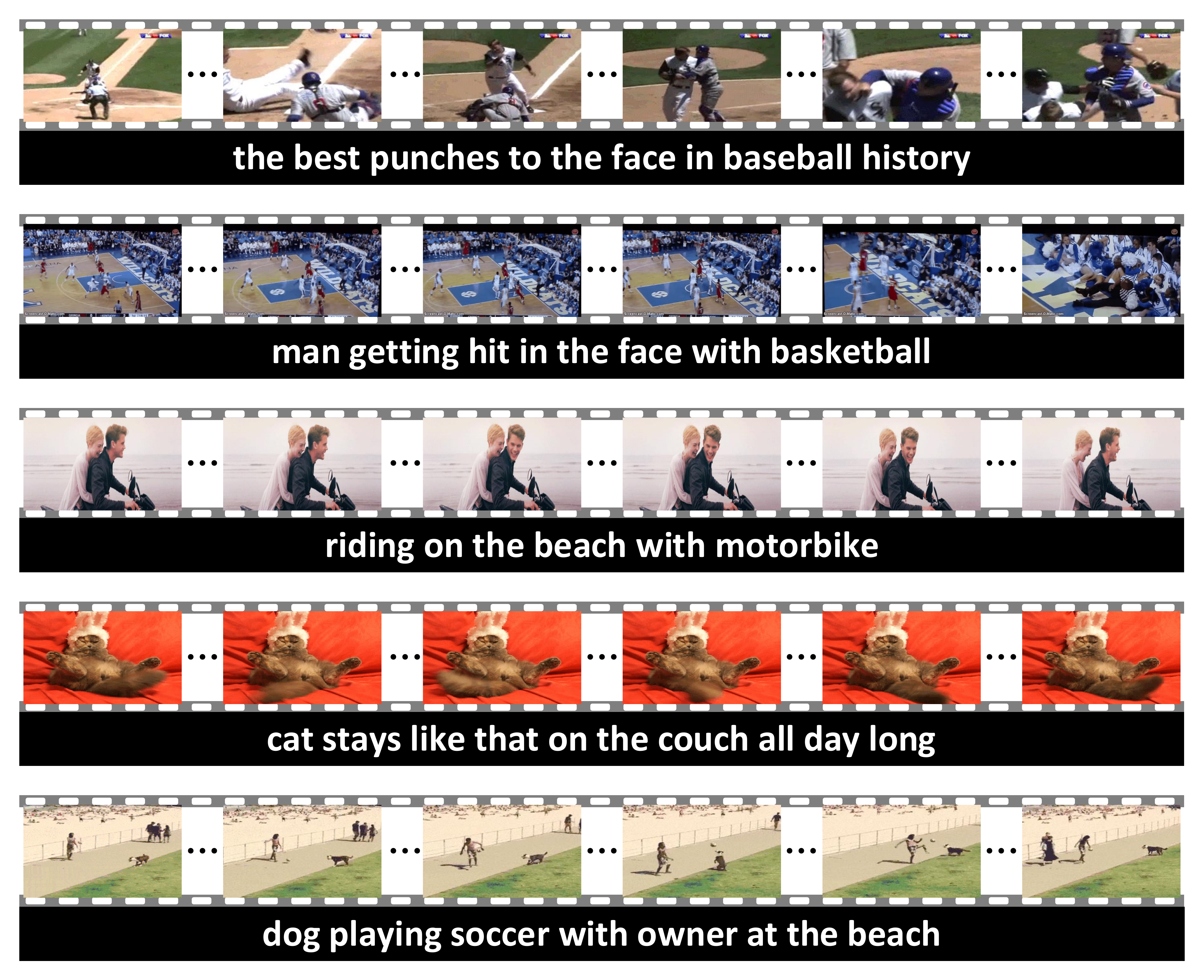}}
\vspace{-0.10in}
\caption{Examples of the GIF videos and the automatically extracted descriptions in our Auto-captions on GIF dataset. We give five samples, with each containing six frames to represent the GIF video and the corresponding sentence.}
\label{fig:figintro}
\vspace{-0.10in}
\end{figure}

With such large-scale programmatically created video-sentence data, we can pre-learn the generic representation or encoder-decoder structure via vision-language pre-training. The pre-trained generic representation or structure can better reflect the cross-modal interaction in a free way and thus benefit a series of downstream video-language tasks, such as video captioning \cite{li2018jointly,pan2017video,Venugopalan:ICCV15}, sentence localization in videos \cite{anne2017localizing}, sentence-to-video generation \cite{pan2017to}, and video question answering \cite{lei2018tvqa}. Technically, we devise a pre-trainable Transformer-based Encoder-Decoder structure (TransED) for vision-language pre-training in video domain. Most specifically, the encoder-decoder structure is first pre-trained on Auto-captions on GIF dataset with four common proxy tasks (masked sequence generation, masked frame-feature regression, video-sentence matching, and masked language modeling). After that, the learnt encoder-decoder structure is further fine-tuned over MSR-VTT for the downstream task of video captioning.

In summary, we make the following contributions in this work:
\textbf{(I).} We build to-date the first automatically generated video-sentence dataset with diverse video content.
\textbf{(II).} We design a Transformer-based encoder-decoder structure for vision-language pre-training in video domain.
\textbf{(III).} We demonstrate the effectiveness of exploiting vision-language pre-training over our Auto-captions on GIF dataset, that facilitates video captioning downstream task.
\section{Auto-captions on GIF Dataset}

The Auto-captions on GIF dataset is characterized by the unique properties including the large-scale video-sentence pairs and the automatic collection process, as well as the comprehensive and diverse video content. In this section, we introduce the automatic pipeline for constructing this dataset in detail, followed by the summarization of our Auto-captions on GIF in comparison to existing video-sentence datasets.

\subsection{Collection of Comprehensive GIF Videos}

Most of existing video-sentence datasets mainly focus on specific fine-grained domains. This adversely hinders the generalization of pre-learnt representation or structure on downstream tasks. For instance, YouCook \cite{das2013thousand:CVPR13} and TACoS \cite{regneri:TACL2013,Rohrbach:ICCV13} are constructed in cooking scenario. MPII-MD \cite{Rohrbach:CVPR15} and M-VAD \cite{Torabi15} focus on movie domain. In order to collect comprehensive and representative GIF videos, we first extract the objects, actions, and SVO (subject-verb-object) triplets from all the sentences in several existing image/video benchmarks (e.g., MSCOCO, MSR-VTT, MSVD, and Conceptual Captions). All the massive extracted items ($\sim$ 1,200,000) are taken as the search queries, and we crawl the GIF videos on web pages via several commercial GIF video search engines for each query. We remove the invalid GIF videos. Ultimately, we collect an original set of comprehensive and representative GIF videos from billions of web pages.

\subsection{Filtering of Sentences}

Next, for each crawled GIF video, we harvest the corresponding raw sentence from the Alt-text HTML attribute. All the raw sentences are filtered as following:

\begin{itemize}
\item We discard the sentences that score too high/low on the polarity annotations via NLTK \cite{loper2002nltk}, or trigger the pornography/profanity detectors \footnote {\small \url{https://pypi.org/project/profanity-filter/};\url{https://pypi.org/project/better-profanity/};\url{https://github.com/areebbeigh/profanityfilter};\url{https://pypi.org/project/profanity-filter/};\url{https://github.com/areebbeigh/profanityfilter}}.
\item The sentences with a high rate of token repetition are filtered out.
\item By parsing sentences via NLTK \cite{loper2002nltk}, we discard the ones with no determiner, no noun, or no preposition.
\item The sentences containing questions, and specific names of movie, TV show, or music video, are discarded.
\item We discard the sentences with the pre-defined high-frequency but less informative phrases (e.g., ``proverb of the day'' and ``this week in rock'').
\item The pre-defined boiler-plate prefix/suffix (e.g., ``click on this'' and ``back to the top of the page link'') in sentences are cropped.
\end{itemize}

\begin{table*}[!tb]
\centering
\caption{Comparison of video-sentence datasets.}
\begin{tabular}{ccccccc}
\Xhline{2\arrayrulewidth}
Dataset   & Context        & Sentence Source & \#Video  & \#Sentence & \#Word    & Vocabulary \\ \hline
YouCook \cite{das2013thousand:CVPR13}  & cooking        & labeled         & -       & 2,668      & 42,457    & 2,711      \\
TACos \cite{regneri:TACL2013,Rohrbach:ICCV13}    & cooking        & AMT workers     & 7,206   & 18,227     & -         & -          \\
TACos M-L \cite{rohrbach:GCPR2014} & cooking        & AMT workers     & 14,105  & 52,593     & -         & -          \\
M-VAD  \cite{Torabi15}   & movie          & DVS             & 48,986  & 55,905     & 519,933   & 18,269     \\
MPII-MD \cite{Rohrbach:CVPR15}  & movie          & DVS+Script      & 68,337  & 68,375     & 653,467   & 24,549     \\
MSVD \cite{Chen:ACL11}     & multi-category & AMT workers     & 1,970   & 70,028     & 607,339   & 13,010     \\
TGIF \cite{tgif:CVPR2016}     & multi-category & Crowd workers   & 102,068 & 125,781    & 1,418,775 & 11,806     \\
MSR-VTT \cite{monkeyxu:CVPR2016}  & 20 categories  & AMT workers     & 10,000  & 200,000    & 1,856,523 & 29,316     \\ \hline
Auto-captions on GIF  & multi-category & Automatic crawling from web             & 163,183 & 164,378    & 1,619,648 & 31,662     \\ \Xhline{2\arrayrulewidth}
\end{tabular}
\label{table:dataset}
\end{table*}

\begin{figure*}[!tb]
\vspace{-0.1in}
\centering {\includegraphics[width=1\linewidth]{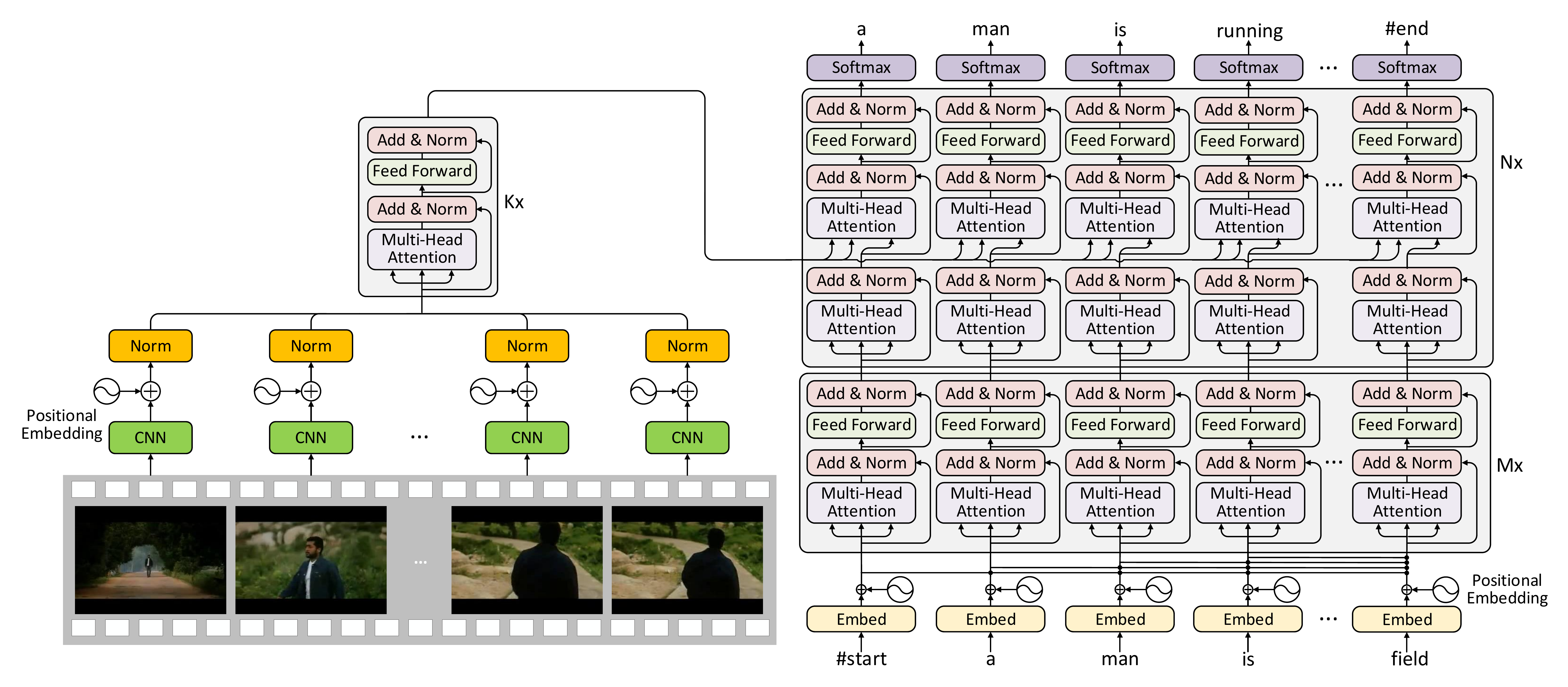}}
\vspace{-0.15in}
\caption{A Transformer-based Encoder-Decoder structure (TransED) for vision-language pre-training, which can be further adapted to the downstream task of video captioning.}
\label{fig:framework}
\vspace{-0.15in}
\end{figure*}

\subsection{Filtering of Video-sentence Pairs}

The previous filtering of sentences stage only examines and filters the raw sentences, leaving the inherent relations between GIF videos and sentences unexploited. Next we additionally filter the video-sentence pairs depending on the semantic relevance in between. In particular, with the assumption that each crawled GIF video is semantically correlated to the search query, we discard the sentence that has no overlap with the search query of the corresponding GIF video. As such, this filtering stage discards the semantically mismatched video-sentence pairs.

\subsection{Selection of Human-like Sentences}

To further screen out the sentences which are similar to human-written descriptions, we train two binary classifiers to recognize whether each sentence is manually written, depending on the whole sentence or the parsed SVO triplet, respectively. Specifically, we take all the human-written sentences in existing image/video captioning benchmarks (e.g., MSCOCO, MSR-VTT, MSVD, and Conceptual Captions) as positive samples, and all the discarded raw sentences in the filtering of sentences stage as negative samples. Finally, only the sentences that simultaneously pass the two classifiers will be taken as the human-like ones for constructing the final dataset.

\subsection{Data Statistics}

Table \ref{table:dataset} details the statistics and comparison among different video-sentence dataset. Note that we are continuing to crawl more GIF videos from new web pages, and thus more data will be released in the future. In current version, our Auto-captions on GIF contains 163,183 GIF videos and 164,378 sentences, and is the largest video-sentence dataset in terms of video number (163,183) and word vocabulary (31,662). Moreover, different from the most existing datasets which focus on specific fine-grained domains and require human annotations, our Auto-captions on GIF is derived from billions of web
pages with massive video categories. As such, the resources can significantly benefit the generalization capability of pre-trained representation or encoder-decoder structure on downstream tasks. To sum up, Auto-captions on GIF represents the most comprehensive, diverse, and complex video-sentence dataset for video understanding, and thus can naturally facilitate the vision-language pre-training in video domain.

\section{Vision-language Pre-training}

Inspired by the recent successes of Transformer self-attention networks \cite{pan2020x,sharma2018conceptual} for vision-language tasks, we present a base model with Transformer-based encoder-decoder structure to access the impact of Auto-captions on GIF dataset for vision-language pre-training.

\textbf{Encoder-Decoder Structure.} Figure \ref{fig:framework} details the architecture of the Transformer-based Encoder-Decoder structure (TransED). Technically, for video encoder, we utilize $K=6$ stacked multi-head self-attention layers to model the self-attention among input frames. The language decoder consists of $M=3$ multi-head self-attention layers and $N=6$ multi-head cross-attention layers (each cross-attention layer is composed of a self-attention sub-layer and a cross-attention sub-layer). More specifically, the stacked multi-head self-attention layers are firstly leveraged to capture the word dependency. Furthermore, the multi-head cross-attention layers are utilized to exploit the co-attention between visual content (frame features from video encoder) and textual tokens (input words).

\textbf{Proxy Tasks for Vision-language Pre-training.} In order to endow the base structure with the capabilities of multi-modal reasoning between vision and language, we pre-train TransED with four vision-language proxy tasks on Auto-captions on GIF dataset: (1) masked language modeling \cite{su2019vl,tan2019lxmert}; (2) masked frame-feature regression as in \cite{tan2019lxmert}; (3) video-sentence matching (in analogy to image-sentence matching \cite{lu2019vilbert}); (4) sequence to sequence generation \cite{zhou2019unified}.

\begin{table}[!tb] \small
\setlength\tabcolsep{4.0 pt}
\centering
\caption{Performance comparisons on MSR-VTT with official split, where B@4, M, R, C and S are short for BLEU@4, METEOR, ROUGE-L, CIDEr-D and SPICE scores. All values are reported as percentage (\%). The short name in the brackets indicates the frame/clip features, where G, C, R, I and A denotes GoogleNet, C3D, ResNet, Inception-Resnet-V2 and Audio feature. }
\begin{tabular}{l|ccccc}
\Xhline{2\arrayrulewidth}
   Model                & B@4  & M    & R    & C    & S   \\ \hline
   MP-LSTM (R) \cite{Venugopalan14}   & 34.1 & 25.4 & - & 35.8 & -   \\
   TA (R) \cite{Yao:ICCV15}                  & 33.2 & 24.9 & -    & 34.5 & -   \\
   S2VT (R) \cite{Venugopalan:ICCV15}        & 34.4 & 25.8 & -    & 36.7 & -   \\
   LSTM-E (R) \cite{pan2016jointly}          & 34.5 & 25.7 & -    & 36.1 & -   \\
   MA-LSTM (G+C+A) \cite{xu2017learning}     & 36.5 & 26.5 & 59.8 & 41.0 & -   \\
   MCNN+MCF (R) \cite{wu2018multi}           & 38.1 & 27.2 & -    & 42.1 & -   \\
   PickNet (R)  \cite{picknet}               & 39.4 & 27.3 & 59.7 & 42.3 & -   \\
   SibNet (G) \cite{liu2018sibnet}           & 40.9 & 27.5 & 60.2 & 47.5 & -   \\
   HRL (R)    \cite{wang2018video}           & 41.3 & 28.7 & 61.7 & 48.0 & -   \\
   TDConvED (R) \cite{chen2019temporal}      & 39.5 & 27.5 & 59.3 & 42.8 & -   \\
   GRU-EVE (I+C) \cite{aafaq2019spatio}      & 36.1 & 27.7 & 59.9 & 45.2 & -   \\
   MARN (R+C)  \cite{pei2019memory}          & 40.4 & 28.1 & 60.7 & 47.1 & -   \\
   MGSA (I+C)  \cite{chen2019motion}         & 42.4 & 27.6 & -    & 47.5 & -   \\
   POS+VCT (R) \cite{hou2019joint}           & 41.4 & 28.9 & 62.0 & 48.1 & -   \\ \hline
   TransED (R)             & 38.3 & 26.8 & 59.2 & 44.3 & 5.8 \\
   TransED+Pre-training (R)    & 39.0 & 27.3 & 59.7 & 45.2 & 5.9 \\
   TransED$_{RL}$ (R)          & 40.2 & 28.3 & 61.0 & 53.6 & 6.8 \\
   TransED$_{RL}$+Pre-training (R) & 41.0 & 28.5 & 61.4 & 54.4 & 6.9 \\
\Xhline{2\arrayrulewidth}
\end{tabular}
\label{table:standard}
\vspace{-0.15in}
\end{table}

\section{Experiments}

In this section, we fully verify the merit of using Auto-captions on GIF for vision-language pre-training and then fine-tuning the pre-trained TransED on MSR-VTT for video captioning downstream task.

\subsection{Datasets and Implementation Details}

\textbf{Pre-training Data of Auto-captions on GIF.} The Auto-captions on GIF contains 163,183 GIF videos and 164,378 sentences, and we utilize the whole dataset for pre-training the base encoder-decoder structure (TransED). For each GIF video, we take all the frames as inputs (maximum frame number: 50).

\textbf{Fine-tuning Data of MSR-VTT.}  MSR-VTT is a widely adopted video-sentence dataset for video captioning task, which consists of 10,000 video clips from 20 well-defined categories. There are 6,513 training videos, 497 validation videos, and 2,990 testing videos in the official split. For the downstream task of video captioning, we fine-tune the pre-trained TransED on the training data of MSR-VTT in the official split. In addition, we also evaluate the pre-trained TransED on the online testing set by submitting the results to online testing server \footnote {\small \url{http://www.auto-video-captions.top/2020/leaderboard}}. For each video in MSR-VTT, we sample the frames at 3 fps and the maximum number of frames is also set as 50. During the fine-tuning stage on MSR-VTT, we optimize TransED with cross-entropy loss. Note that we involve a variant of TransED (named TransED$_{RL}$) which is further optimized with CIDEr reward.

\begin{table}[!tb] \small
\setlength\tabcolsep{3 pt}
\centering
\caption{Performance comparisons on online testing server.}
\begin{tabular}{l|ccccc}
\Xhline{2\arrayrulewidth}
Model                 & B@4  & M    & R    & C    & S   \\ \hline
\multicolumn{6}{l}{\emph{Fine-tune with 6.5k videos (train split), online evaluation}}               \\ \hline
TransED (R)             & 16.4 & 15.5 & 39.1 & 17.0 & 4.4 \\
TransED+Pre-training (R)    & 17.1 & 15.8 & 39.5 & 18.0 & 4.6 \\
TransED$_{RL}$ (R)          & 16.6 & 15.8 & 40.0 & 20.4 & 4.8 \\
TransED$_{RL}$+Pre-training (R) & 18.1 & 16.4 & 40.9 & 22.3 & 5.1 \\ \hline
\multicolumn{6}{l}{\emph{Fine-tune with 9.5k videos (train+test splits), online evaluation}}              \\ \hline
TransED (R)             & 17.4 & 16.2 & 39.6 & 19.6 & 4.8 \\
TransED+Pre-training (R)    & 18.8 & 16.3 & 40.6 & 19.7 & 4.8 \\
TransED$_{RL}$ (R)          & 17.9 & 16.3 & 40.5 & 22.5 & 5.1 \\
TransED$_{RL}$+Pre-training (R) & 19.5 & 16.8 & 41.3 & 23.9 & 5.4 \\
\Xhline{2\arrayrulewidth}
\end{tabular}
\label{table:online}
\vspace{-0.15in}
\end{table}

\subsection{Performance Comparison}

\textbf{Offline Evaluation on Official Split.}
Table \ref{table:standard} shows the performance comparisons on MSR-VTT with official split. It is worth noting that the reported performances of different state-of-the-art task-specific models are often based on different frame/clip representations. For fair comparisons, we evaluate our base models (TransED, TransED$_{RL}$) on the most commonly adopted frame representation (i.e., the output from ResNet). Moreover, we involve two different experimental settings for each base model: TransED/TransED$_{RL}$ denotes the base model which is only trained with task-specific data, without pre-training on our Auto-captions on GIF dataset; TransED/TransED$_{RL}$+Pre-training represents that the base model is pre-trained over Auto-captions on GIF and further fine-tuned on task-specific data.

Overall, under the same task-specific setting without vision-language pre-training, TransED and TransED$_{RL}$ obtain comparable results with other state-of-the-art task-specific models. Furthermore, by pre-training TransED/TransED$_{RL}$ on Auto-captions on GIF and then fine-tuning it on MSR-VTT, the TransED/TransED$_{RL}$+Pre-training consistently exhibits better performances than TransED/TransED$_{RL}$ across all the evaluation metrics. This confirms the merit of exploiting vision-language pre-training over our Auto-captions on GIF, that facilitates the downstream task of video captioning on MSR-VTT.

\textbf{Online Evaluation on Online Testing Server.}
In addition, we evaluate the base models on the online testing set. Table \ref{table:online} details the performances over online testing videos. Note that here we adopt two different sets (6.5k training videos, and 9.5k training plus testing videos in official split) for fine-tuning TransED/TransED$_{RL}$ on MSR-VTT. Similar to the observations in offline evaluation, TransED/TransED$_{RL}$+Pre-training performs better than TransED/TransED$_{RL}$ by additionally pre-training the based model on Auto-captions on GIF.

\section{Conclusions}
We introduced a new video-sentence dataset, Auto-captions on GIF, which is automatically created from billions of web pages. This dataset contains to-date the largest amount of videos with the most comprehensive and representative video content, and thus supports vision-language pre-training in video domain. We experimentally evaluated the base models with Transformer-based encoder-decoder structure for vision-language pre-training over our Auto-captions on GIF dataset. The results demonstrate the compelling generalizability of pre-trained encoder-decoder structure by fine-tuning it to video captioning downstream task on MSR-VTT.

{\small
\bibliographystyle{ieee_fullname}
\bibliography{egbib}

\begin{thebibliography}{10}\itemsep=-1pt

\bibitem{aafaq2019spatio}
Nayyer Aafaq, Naveed Akhtar, Wei Liu, Syed~Zulqarnain Gilani, and Ajmal Mian.
\newblock Spatio-temporal dynamics and semantic attribute enriched visual
  encoding for video captioning.
\newblock In {\em CVPR}, 2019.

\bibitem{anderson2017bottom}
Peter Anderson, Xiaodong He, Chris Buehler, Damien Teney, Mark Johnson, Stephen
  Gould, and Lei Zhang.
\newblock Bottom-up and top-down attention for image captioning and visual
  question answering.
\newblock In {\em CVPR}, 2018.

\bibitem{anne2017localizing}
Lisa Anne~Hendricks, Oliver Wang, Eli Shechtman, Josef Sivic, Trevor Darrell,
  and Bryan Russell.
\newblock Localizing moments in video with natural language.
\newblock In {\em ICCV}, 2017.

\bibitem{antol2015vqa}
Stanislaw Antol, Aishwarya Agrawal, Jiasen Lu, Margaret Mitchell, Dhruv Batra,
  C Lawrence~Zitnick, and Devi Parikh.
\newblock Vqa: Visual question answering.
\newblock In {\em ICCV}, 2015.

\bibitem{Chen:ACL11}
David~L Chen and William~B Dolan.
\newblock Collecting highly parallel data for paraphrase evaluation.
\newblock In {\em ACL}, 2011.

\bibitem{chen2019temporal}
Jingwen Chen, Yingwei Pan, Yehao Li, Ting Yao, Hongyang Chao, and Tao Mei.
\newblock Temporal deformable convolutional encoder-decoder networks for video
  captioning.
\newblock In {\em AAAI}, 2019.

\bibitem{chen2019motion}
Shaoxiang Chen and Yu-Gang Jiang.
\newblock Motion guided spatial attention for video captioning.
\newblock In {\em AAAI}, 2019.

\bibitem{picknet}
Yangyu Chen, Shuhui Wang, Weigang Zhang, and Qingming Huang.
\newblock Less is more: Picking informative frames for video captioning.
\newblock In {\em ECCV}, 2018.

\bibitem{das2013thousand:CVPR13}
Pradipto Das, Chenliang Xu, Richard~F Doell, and Jason~J Corso.
\newblock A thousand frames in just a few words: Lingual description of videos
  through latent topics and sparse object stitching.
\newblock In {\em CVPR}, 2013.

\bibitem{hou2019joint}
Jingyi Hou, Xinxiao Wu, Wentian Zhao, Jiebo Luo, and Yunde Jia.
\newblock Joint syntax representation learning and visual cue translation for
  video captioning.
\newblock In {\em ICCV}, 2019.

\bibitem{lei2018tvqa}
Jie Lei, Licheng Yu, Mohit Bansal, and Tamara Berg.
\newblock Tvqa: Localized, compositional video question answering.
\newblock In {\em EMNLP}, 2018.

\bibitem{li2019learning}
Yehao Li, Yingwei Pan, Ting Yao, Hongyang Chao, Yong Rui, and Tao Mei.
\newblock Learning click-based deep structure-preserving embeddings with visual
  attention.
\newblock {\em ACM Transactions on Multimedia Computing, Communications, and
  Applications (TOMM)}, 2019.

\bibitem{tgif:CVPR2016}
Yuncheng Li, Yale Song, Liangliang Cao, Joel Tetreault, Larry Goldberg,
  Alejandro Jaimes, and Jiebo Luo.
\newblock Tgif: A new dataset and benchmark on animated gif description.
\newblock In {\em CVPR}, 2016.

\bibitem{li2018jointly}
Yehao Li, Ting Yao, Yingwei Pan, Hongyang Chao, and Tao Mei.
\newblock Jointly localizing and describing events for dense video captioning.
\newblock In {\em CVPR}, 2018.

\bibitem{li2019novel}
Yehao Li, Ting Yao, Yingwei Pan, Hongyang Chao, and Tao Mei.
\newblock Pointing novel objects in image captioning.
\newblock In {\em CVPR}, 2019.

\bibitem{liu2018sibnet}
Sheng Liu, Zhou Ren, and Junsong Yuan.
\newblock Sibnet: Sibling convolutional encoder for video captioning.
\newblock In {\em ACM MM}, 2018.

\bibitem{loper2002nltk}
Edward Loper and Steven Bird.
\newblock Nltk: The natural language toolkit.
\newblock In {\em Proceedings of the ACL-02 Workshop on Effective Tools and
  Methodologies for Teaching Natural Language Processing and Computational
  Linguistics}, 2002.

\bibitem{lu2019vilbert}
Jiasen Lu, Dhruv Batra, Devi Parikh, and Stefan Lee.
\newblock Vilbert: Pretraining task-agnostic visiolinguistic representations
  for vision-and-language tasks.
\newblock In {\em NeurIPS}, 2019.

\bibitem{pan2016jointly}
Yingwei Pan, Tao Mei, Ting Yao, Houqiang Li, and Yong Rui.
\newblock Jointly modeling embedding and translation to bridge video and
  language.
\newblock In {\em CVPR}, 2016.

\bibitem{pan2017to}
Yingwei Pan, Zhaofan Qiu, Ting Yao, Houqiang Li, and Tao Mei.
\newblock To create what you tell: Generating videos from captions.
\newblock In {\em MM}, 2017.

\bibitem{pan2017video}
Yingwei Pan, Ting Yao, Houqiang Li, and Tao Mei.
\newblock Video captioning with transferred semantic attributes.
\newblock In {\em CVPR}, 2017.

\bibitem{pan2020x}
Yingwei Pan, Ting Yao, Yehao Li, and Tao Mei.
\newblock X-linear attention networks for image captioning.
\newblock In {\em CVPR}, 2020.

\bibitem{pan2014click}
Yingwei Pan, Ting Yao, Tao Mei, Houqiang Li, Chong-Wah Ngo, and Yong Rui.
\newblock Click-through-based cross-view learning for image search.
\newblock In {\em SIGIR}, 2014.

\bibitem{pei2019memory}
Wenjie Pei, Jiyuan Zhang, Xiangrong Wang, Lei Ke, Xiaoyong Shen, and Yu-Wing
  Tai.
\newblock Memory-attended recurrent network for video captioning.
\newblock In {\em CVPR}, 2019.

\bibitem{regneri:TACL2013}
Michaela Regneri, Marcus Rohrbach, Dominikus Wetzel, Stefan Thater, Bernt
  Schiele, and Manfred Pinkal.
\newblock Grounding action descriptions in videos.
\newblock {\em Transactions of the Association for Computational Linguistics},
  1:25--36, 2013.

\bibitem{rohrbach:GCPR2014}
Anna Rohrbach, Marcus Rohrbach, Wei Qiu, Annemarie Friedrich, Manfred Pinkal,
  and Bernt Schiele.
\newblock Coherent multi-sentence video description with variable level of
  detail.
\newblock In {\em German conference on pattern recognition}, pages 184--195,
  2014.

\bibitem{Rohrbach:CVPR15}
Anna Rohrbach, Marcus Rohrbach, Niket Tandon, and Bernt Schiele.
\newblock A dataset for movie description.
\newblock In {\em CVPR}, 2015.

\bibitem{Rohrbach:ICCV13}
Marcus Rohrbach, Wei Qiu, Ivan Titov, Stefan Thater, Manfred Pinkal, and Bernt
  Schiele.
\newblock Translating video content to natural language descriptions.
\newblock In {\em ICCV}, 2013.

\bibitem{sharma2018conceptual}
Piyush Sharma, Nan Ding, Sebastian Goodman, and Radu Soricut.
\newblock Conceptual captions: A cleaned, hypernymed, image alt-text dataset
  for automatic image captioning.
\newblock In {\em ACL}, 2018.

\bibitem{su2019vl}
Weijie Su, Xizhou Zhu, Yue Cao, Bin Li, Lewei Lu, Furu Wei, and Jifeng Dai.
\newblock Vl-bert: Pre-training of generic visual-linguistic representations.
\newblock {\em arXiv preprint arXiv:1908.08530}, 2019.

\bibitem{tan2019lxmert}
Hao Tan and Mohit Bansal.
\newblock Lxmert: Learning cross-modality encoder representations from
  transformers.
\newblock In {\em EMNLP-IJCNLP}, 2019.

\bibitem{Torabi15}
Atousa Torabi, Christopher Pal, Hugo Larochelle, and Aaron Courville.
\newblock Using descriptive video services to create a large data source for
  video annotation research.
\newblock {\em arXiv preprint arXiv:1503.01070}, 2015.

\bibitem{Venugopalan:ICCV15}
Subhashini Venugopalan, Marcus Rohrbach, Jeffrey Donahue, Raymond Mooney,
  Trevor Darrell, and Kate Saenko.
\newblock Sequence to sequence - video to text.
\newblock In {\em ICCV}, 2015.

\bibitem{Venugopalan14}
Subhashini Venugopalan, Huijuan Xu, Jeff Donahue, Marcus Rohrbach, Raymond
  Mooney, and Kate Saenko.
\newblock Translating videos to natural language using deep recurrent neural
  networks.
\newblock In {\em NAACL HLT}, 2015.

\bibitem{wang2019paragraph}
Jing Wang, Yingwei Pan, Ting Yao, Jinhui Tang, and Tao Mei.
\newblock Convolutional auto-encoding of sentence topics for image paragraph
  generation.
\newblock In {\em IJCAI}, 2019.

\bibitem{wang2018video}
Xin Wang, Wenhu Chen, Jiawei Wu, Yuan-Fang Wang, and William Yang~Wang.
\newblock Video captioning via hierarchical reinforcement learning.
\newblock In {\em CVPR}, 2018.

\bibitem{wu2018multi}
Aming Wu and Yahong Han.
\newblock Multi-modal circulant fusion for video-to-language and backward.
\newblock In {\em IJCAI}, 2018.

\bibitem{monkeyxu:CVPR2016}
Jun Xu, Tao Mei, Ting Yao, and Yong Rui.
\newblock Msr-vtt: A large video description dataset for bridging video and
  language.
\newblock In {\em CVPR}, 2016.

\bibitem{xu2017learning}
Jun Xu, Ting Yao, Yongdong Zhang, and Tao Mei.
\newblock Learning multimodal attention lstm networks for video captioning.
\newblock In {\em ACM MM}, 2017.

\bibitem{Yao:ICCV15}
Li Yao, Atousa Torabi, Kyunghyun Cho, Nicolas Ballas, Christopher Pal, Hugo
  Larochelle, and Aaron Courville.
\newblock Describing videos by exploiting temporal structure.
\newblock In {\em ICCV}, 2015.

\bibitem{yao2015learning}
Ting Yao, Tao Mei, and Chong-Wah Ngo.
\newblock Learning query and image similarities with ranking canonical
  correlation analysis.
\newblock In {\em ICCV}, 2015.

\bibitem{yao2017novel}
Ting Yao, Yingwei Pan, Yehao Li, and Tao Mei.
\newblock Incorporating copying mechanism in image captioning for learning
  novel objects.
\newblock In {\em CVPR}, 2017.

\bibitem{yao2018exploring}
Ting Yao, Yingwei Pan, Yehao Li, and Tao Mei.
\newblock Exploring visual relationship for image captioning.
\newblock In {\em ECCV}, 2018.

\bibitem{yao2019hierarchy}
Ting Yao, Yingwei Pan, Yehao Li, and Tao Mei.
\newblock Hierarchy parsing for image captioning.
\newblock In {\em ICCV}, 2019.

\bibitem{yao2017boosting}
Ting Yao, Yingwei Pan, Yehao Li, Zhaofan Qiu, and Tao Mei.
\newblock Boosting image captioning with attributes.
\newblock In {\em ICCV}, 2017.

\bibitem{zhou2019unified}
Luowei Zhou, Hamid Palangi, Lei Zhang, Houdong Hu, Jason~J Corso, and Jianfeng
  Gao.
\newblock Unified vision-language pre-training for image captioning and vqa.
\newblock In {\em AAAI}, 2020.

\end{thebibliography}
}

\end{document}